\documentclass[letterpaper, 10 pt, conference]{ieeeconf}  

\IEEEoverridecommandlockouts                              

\usepackage{graphicx} 
\usepackage{amsmath} 
\usepackage{amssymb}  
\usepackage{xcolor}
\usepackage{hyperref}
\usepackage{balance}
\usepackage{cite}
\usepackage{booktabs}
\usepackage{multirow}
\usepackage{array}

\DeclareMathOperator*{\argmin}{arg\,min}

\title{\LARGE \bf
Stable Tracking-in-the-Loop Control of Cable-Driven Surgical Manipulators under Erroneous Kinematic Chains
}

\author{Neelay Joglekar$^{1}$, Fei Liu$^{2}$, Florian Richter$^{1}$, and Michael C. Yip$^{1}$%
\thanks{This work was supported by NIH 1-R21-EB036284 and NSF Award 2045803}%
\thanks{$^{1}$ Department of Electrical and Computer Engineering, University of California San Diego, La Jolla, CA 92093, USA {\tt\small \{njogleka, yip\}@ucsd.edu, frichter1995@gmail.com}.}%
\thanks{$^2$ Department of Electrical Engineering and Computer Science, University of Tennessee Knoxville, Knoxville, TN 37996, USA {\tt\small fliu33@utk.edu}}%
}

\begin{document}
\setlength{\tabcolsep}{3pt}

\maketitle
\thispagestyle{empty}
\pagestyle{empty}

\begin{abstract}

Remote Center of Motion (RCM) robotic manipulators have revolutionized Minimally Invasive Surgery, enabling precise, dexterous surgical manipulation within the patient's body cavity without disturbing the insertion point on the patient. Accurate RCM tool control is vital for incorporating autonomous subtasks like suturing, blood suction, and tumor resection into robotic surgical procedures, reducing surgeon fatigue and improving patient outcomes. However, these cable-driven systems are subject to significant joint reading errors, corrupting the kinematics computation necessary to perform control. Although visual tracking with endoscopic cameras can correct errors on in-view joints, errors in the kinematic chain prior to the insertion point are irreparable because they remain out of view. No prior work has characterized the stability of control under these conditions. We fill this gap by designing a provably stable tracking-in-the-loop controller for the out-of-view portion of the RCM manipulator kinematic chain. We additionally incorporate this controller into a bilevel control scheme for the full kinematic chain. We rigorously benchmark our method in simulated and real world settings to verify our theoretical findings. Our work provides key insights into the next steps required for the transition from teleoperated to autonomous surgery.

\end{abstract}

\section{INTRODUCTION}

Robotic surgical systems provide surgeons more precise, tremor-free control of surgical tools, enabling numerous Robotically Assisted Minimally Invasive Surgery (RAMIS) procedures that improve patient outcomes. This is particularly due to their use of Remote Center of Motion (RCM) manipulators \cite{zhang2024state}, which can freely move about the surgical workspace while remaining stationary at an insertion point. Although current RAMIS procedures are fully teleoperated, automating surgical subtasks such as suturing \cite{ostrander2024current, hari2024stitch}, tumor resection \cite{ge2023autonomous, liang2024medic}, tissue grasping \cite{Fei_2021_R2S}, and blood suction \cite{richter2021autonomous, Jingbin_2021_blood, ou2024autonomous} can reduce surgeon workload and enable them to focus on higher-level decision making. Many prior works have proposed and demonstrated autonomous solutions for these subtasks, but a key roadblock to their implementation has been lack of confidence in their reliability. As a result, there has been growing interest in developing reliability-driven autonomous algorithms \cite{shinde2024surestep, joglekar2024autonomous, shinde2024jiggle} that can confidently handle any uncertainty or risk that may arise in surgical situations.

\begin{figure}[t]
    \centering
    \includegraphics[width=\linewidth]{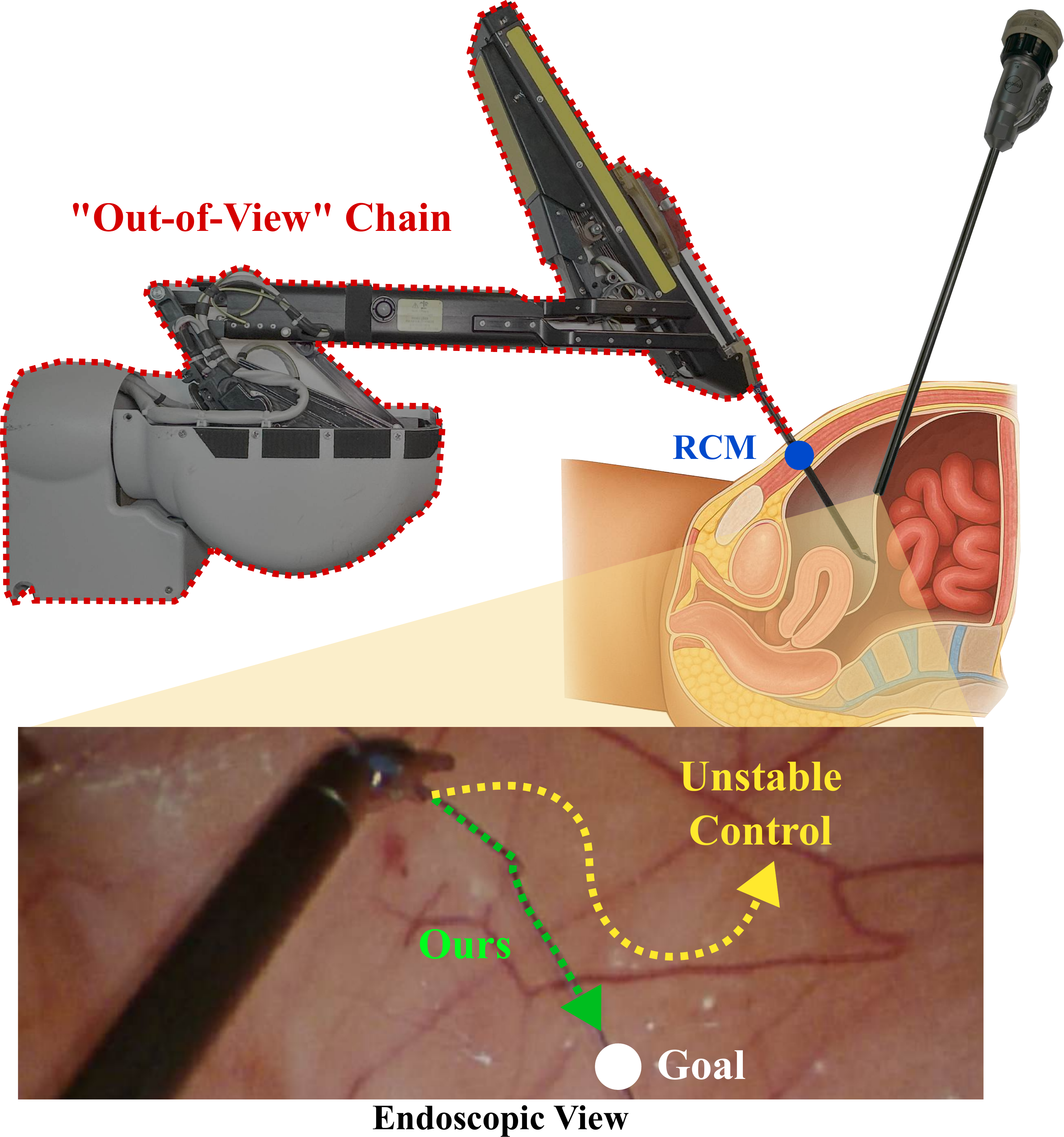}
    \caption{Cable-Driven Remote Center of Motion (RCM) surgical manipulators are prone to kinematic chain errors, and these cannot be fully corrected because a large portion of the robot remains out of view of the endoscopic camera. This can produce unstable control behavior.
    We are the first to provide theoretical analysis on control stability under errors in the RCM kinematic chain, formulating an associated controller with strong stability guarantees. (Torso image altered from \cite{blausen2014medical} with GPT-4o and DALLE to remove annotations).}
    \label{title}
\end{figure}

Enforcing reliable manipulator control is a key prerequisite in ensuring the safety of any autonomous procedure, but this is not straightforward for serial RCM surgical systems. A typical kinematics-based closed-loop controller assumes access to accurate ground-truth joint angle readings. However, in cable-driven RCM manipulators, joint angle measurements are often erroneous due to cable effects such as hysteresis, friction, and backlash. Adding visual tracking into the loop can help correct joint reading errors in observable joints, but several RCM manipulator joints remain out-of-view, as shown in Fig. \ref{title}, and cannot be individually corrected \cite{richter2021robotic}. This produces significant error in the kinematic chain, as shown in Fig. \ref{background}. How does this untracked kinematic error contribute to errors in trajectory following and, under visual closed-loop control, can it produce an unstable control system? No prior work has sought to answer this question, or provide theoretical guarantees on control stability under these conditions, jeapordizing the reliability of RCM manipulator control algorithms.

In this manuscript, we seek to fill this gap, providing the following contributions:
\begin{itemize}
    \item  The first theoretical analysis of tracking-in-the-loop RCM manipulator control stability under kinematic chain inaccuracy;
    \item A resolved-rate controller for the out-of-view component of the RCM kinematic chain that is theoretically stable within a wide joint error bound;
    \item A bilevel control scheme that incorporates our stable out-of-view chain controller into a full chain control algorithm.
\end{itemize}
We validate our results in simulation and hardware experiments, using the da Vinci Research Kit (dVRK) as a case study. Our work provides crucial insight into the next steps required to the improve the reliability of surgical autonomy.

\section{RELATED WORKS}

Initial works used sophisticated calibration techniques to account for cable effects in RCM surgical tool control \cite{peng2020real, hwang2020efficiently}. For example, Hwang et al. used fiducial attachments to collect a large dataset of tool poses and joint angles, which was used to train an RNN-based controller. However, calibrations degrade over time through cable creep, where cables irreversibly stretch over time.

To address this, other works incorporate visual tracking into the loop. Although it would be ideal to track camera extrinsic calibration error and each joint angle error as a separate term, Richter et al. proved it's typically impossible to estimate these quantities individually \cite{richter2021robotic}. Instead, most works track the combined effect of these errors \cite{ye2016real, hao2018vision}, often referred to either as the ``kinematics remote-center coordinate system (KCS)" \cite{reiter2014appearance, zhao2015efficient} or the ``lumped" error correction term \cite{richter2021robotic, dambrosia2024robust}.

\begin{figure}[t]
    \centering
    \vspace{5mm}\includegraphics[width=\linewidth]{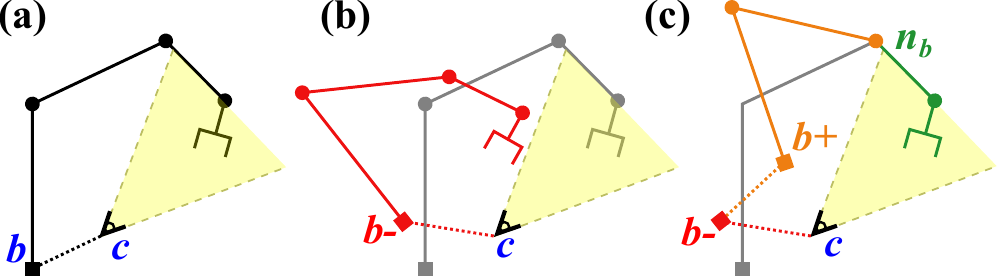}
    \caption{A 2D example: a) An ideal kinematic chain can be constructed from joint readings and a base-to-camera transform. b) However, camera calibration and joint reading errors produce an erroneous kinematic chain (red). c) ``Lumped error" tracking paired with visual joint angle correction repairs the configuration of the in-view chain (green), but the out-of-view chain (orange) remains inaccurate. This kinematic inaccuracy complicates control, especially since the ``lumped error" term (orange dashed) dynamically changes with the joint angles.}
    \label{background}
\end{figure}

Although this approach enables accurate tool tip tracking, it produces a false estimate of the full kinematic chain, as shown in Fig. \ref{background}. Prior works naively relied on this false chain to perform control without providing guarantees on stability \cite{richter2021autonomous, chiu2021bimanual, ho2024surgirl}. As will be shown in Section V, it is indeed possible for visual-servoing based controllers to become unstable when these kinematic chain errors are not properly handled. We seek to fill this knowledge gap in this manuscript, providing insight into how lumped error tracking alters the estimated kinematic chain and performing theoretical analysis on the stability of controlling this chain.

\section{BACKGROUND}

\subsection{Lumped Error Tracking}

Visualized in Fig. \ref{background}(a), the kinematic chain for an ideal manipulator is as follows.
\begin{equation} \label{ideal_kc}
    \mathbf{T} = \mathbf{T}^{c}_{b} \prod_{i=1}^{n} \mathbf{T}^{i-1}_{i} (q_i)
\end{equation}
where $\mathbf{q} = [q_1 \dots q_n]^{\top}$ are the true joint angles, $\mathbf{T}^{i-1}_{i} (q_i)$ is the transform from link $i$ to $i-1$, and $\mathbf{T}^{c}_{b}$ is the base-to-camera transform.

However, as shown in Fig. \ref{background}(b), surgical manipulators typically deal with two types of errors: errors in the base-to-camera calibration and errors in the joint angle readings. In this case, the true chain can be rewritten as follows:
\begin{equation} \label{err_kc}
    \mathbf{T} = \mathbf{T}^{c}_{b-} \mathbf{T}^{b-}_{b}  \prod_{i=1}^{n} \mathbf{T}^{i-1}_{i} (\tilde{q}_i + e_i)
\end{equation}
Here, $\mathbf{\tilde{q}} = [\tilde{q}_1 \dots \tilde{q}_n]^{\top}$ are joint angle readings with errors $\mathbf{e} = [e_1 \dots e_n]^{\top}$, such that $\mathbf{q} = \mathbf{\tilde{q}} + \mathbf{e}$, and 
$\mathbf{T}^{c}_{b-}$ is the base-to-camera transform combined with the camera extrinsic calibration error $\mathbf{T}^{b-}_{b}$.

Naively, we can use visual feedback and attempt to identify $\mathbf{e}$ and $\mathbf{T}^{b-}_{b}$ to compute \eqref{err_kc}. If $n_b$ is the first visible link, then the errors for the in-view joints $\mathbf{e_{n_b+1:n}}$ can be directly estimated from the camera view. However, because the remainder of the chain lies out of the camera view, errors $\mathbf{e_{1:n_b}}$ and $\mathbf{T}^{b-}_{b}$ often cannot be individually reconstructed \cite{richter2021robotic}.

Nevertheless, we can use the camera's view of the end-effector pose to track the \textit{combined} influence of $\mathbf{e_{1:n_b}}$ and $\mathbf{T}^{b-}_{b}$, applied as a ``lumped error" correction $\mathbf{T}^{b-}_{b+}$ to the base-to-camera transform \cite{richter2021robotic}. This produces the following imaginary kinematic chain, visualized in Fig. \ref{background}(c):
\begin{equation} \label{og_kc}
    \mathbf{T} = \mathbf{T}^{c}_{b-} \mathbf{T}^{b-}_{b+} \prod_{i=1}^{n_b} \mathbf{T}^{i-1}_{i} (\tilde{q}^i) \prod_{i=n_b+1}^{n} \mathbf{T}^{i-1}_{i} (\tilde{q}^i + e^i)
\end{equation}
We define $b+$ as the imaginary base frame made by the lumped error transform. We also note that $\mathbf{T}^{b-}_{b+}$ is dependent on $\mathbf{\tilde{q}}$ \cite{richter2021robotic}, so it will change as the robot moves even if $\mathbf{e_{1:n_b}}$ and $\mathbf{T}^{b-}_{b}$ remain constant. Although this formulation enables tracking of the in-view portion of the kinematic chain, the out-of-view portion remains erroneous. This poses a problem, as typical control approaches require not just accurate tracking but also accurate joint angles for each individual joint in the kinematic chain.


\subsection{Inverse Kinematics-Based Control}

To perform control under the partially inaccurate chain described in \eqref{og_kc}, we previously developed a controller that transforms a given goal pose from frame $c$ to frame $b+$ and then solves the inverse kinematics (IK) to achieve that goal. Note that because $\mathbf{T}^{b-}_{b+}$ changes as the robot moves, the controller repeatedly re-solves the IK problem after each action until the end-effector reaches the goal point.

Empirically, this controller works well for several tasks, including suture needle regrasping, suture thread grasping, and blood suction. However, it's difficult to characterize its stability analytically because this would require solving the highly nonlinear system of equations defined by \eqref{og_kc}, which is typically solved numerically. This lack of theoretical backing makes it difficult to gauge the reliability of the IK controller in settings beyond those previously tested.

In this manuscript, we propose a controller with theoretical guarantees on stability within a bounded set of joint angle errors. Furthermore, we show in Section V that this error bound empirically applies to the IK controller as well under certain conditions.

\section{METHODS}

\begin{figure*}[t]
    \centering
    \vspace{5mm}\includegraphics[width=\linewidth]{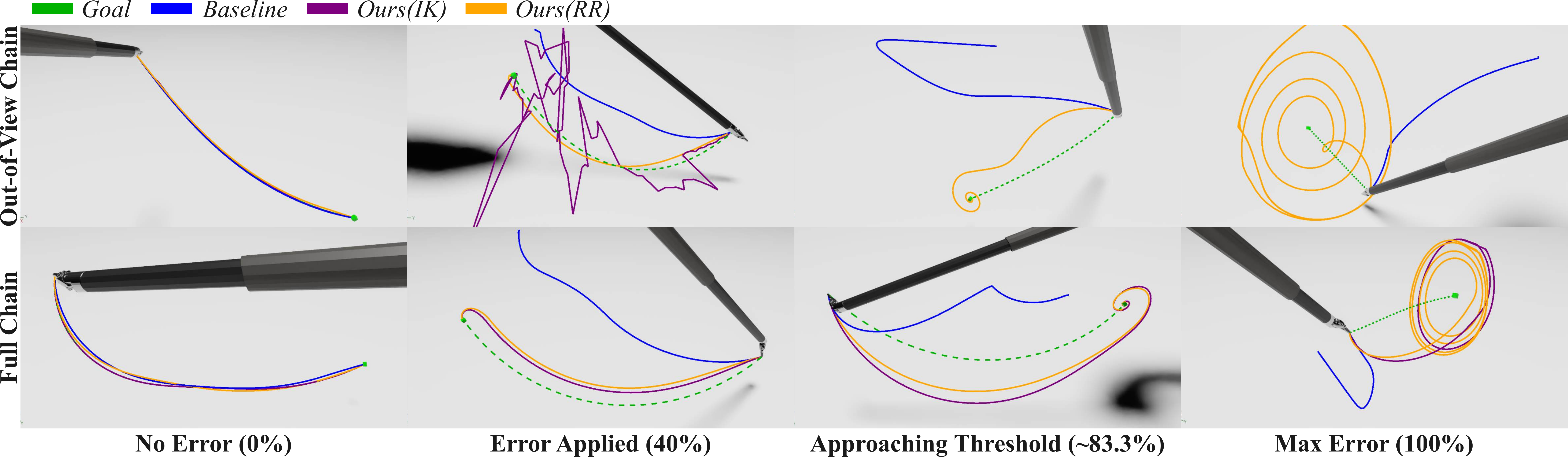}
    \caption{Simulated experiment examples for out-of-view chain (top) and full chain (bottom) control. We gradually added errors to the out-of-view joints and visualized how control performance degraded. The calibration-based baseline rapidly fails in both chain settings. Our IK controller rapidly becomes unstable when controlling the out-of-view chain because it's underactuated, producing a workspace mismatch. However, it closely matches our resolved-rate (RR) controller behavior in the full chain case. In both cases, our resolved-rate controller reliably converges until the joint error exceeds our theoretical threshold, when it begins oscillating around the goal.}
    \label{sim_examples}
\end{figure*}

We specifically focus on serial RCM robotic systems similar to the da Vinci Research Kit (dVRK). Our coordinate frame definitions follow the Modified Denavit-Hartenberg (DH) convention, and for the dVRK the modified DH parameters can be found in \cite{hwang2020efficiently}. In our case $n_b = 4$, so the out-of-view chain consists of the first 4 joints: outer yaw, outer pitch, tool insertion, and tool roll. The in-view chain consists of the remaining distal tool chain which varies depending on tool selection, so we ensure our controller is agnostic to the tool type.

We initially focus on defining a stable controller for the out-of-view chain only, and afterwards integrate this into a bilevel control scheme for full chain control. Our method is a modified version of canonical resolved-rate (RR) motion control \cite{wampler1988applications}, which involves using the kinematic chain Jacobian's pseudoinverse to execute goal task space velocities.

\subsection{Out-of-View Chain Control}

To develop a resolved-rate controller for the out-of-view chain, we initially examine the out-of-view chain's body frame Jacobian (i.e. output twist vectors are in the link $n_b$ frame):
\begin{equation} \label{Jnb}
    \mathbf{J}_{n_b} = \begin{bmatrix}
        -q_3\cos q_2\sin q_4  & q_3\cos q_4  & 0 & 0 \\
        -q_3\cos q_2 \cos q_4  & -q_3\sin q_4  & 0 & 0 \\
        0 & 0 & 1 & 0 \\
        \cos q_2 \cos q_4  & \sin q_4  & 0 & 0 \\
        -\cos q_2 \sin q_4  & \cos q_4  & 0 & 0 \\
        \sin q_2  & 0 & 0 & 1
    \end{bmatrix}
\end{equation}

We can decompose \eqref{Jnb} as follows:
\begin{equation} \label{D}
    \mathbf{D} = \begin{bmatrix}
        1 & 0 & 0 & 0\\
        0 & 1 & 0 & 0\\
        0 & 0 & 1 & 0\\
        0 & \frac{-1}{q_3} & 0 & 0\\
        \frac{1}{q_3} & 0 & 0 & 0\\
        0 & 0 & 0 & 1
    \end{bmatrix}
\end{equation}
\begin{equation}\label{Q}
    \mathbf{Q} = \begin{bmatrix}
        -\sin q_4  & \cos q_4  & 0 & 0 \\
        -\cos q_4  & -\sin q_4  & 0 & 0 \\
        0 & 0 & 1 & 0 \\
        0 & 0 & 0 & 1
    \end{bmatrix}
\end{equation}
\vspace{2mm}
\begin{equation}\label{W}
    \mathbf{W} = \begin{bmatrix}
        q_3\cos q_2  & 0 & 0 & 0\\
        0 & q_3 & 0 & 0\\
        0 & 0 & 1 & 0\\
        0 & 0 & 0 & 1
    \end{bmatrix}
\end{equation}
\begin{equation}\label{S}
    \mathbf{S} = \begin{bmatrix}
        1 & 0 & 0 & 0\\
        0 & 1 & 0 & 0\\
        0 & 0 & 1 & 0 \\
        \sin q_2 & 0 & 0 & 1
    \end{bmatrix}
\end{equation}
\begin{equation}\label{Jnb_decomp}
    \mathbf{J}_{n_b} = \mathbf{D}\mathbf{Q}\mathbf{W}\mathbf{S}
\end{equation}

Intuitively, $\mathbf{Q}, \mathbf{W}, \mathbf{S} \in \mathbb{R}^{4\times4}$ can be thought of as affine operations in a 4D ``linear+roll" space, composed of 3D linear velocity concatenated with the 1D angular velocity about the $z$-axis of link $n_b$. $\mathbf{D} \in \mathbb{R}^{6\times4}$ deprojects these operations into 6D twist space. This decomposition arises because the out-of-view chain has only 4 degrees of freedom.

We can take advantage of this decomposition by defining our control law in the 4D ``linear+roll" subspace rather than 6D twist space. However, we're still unable to properly calculate $\mathbf{Q}, \mathbf{W}, \mathbf{S}$ because we don't have access to $\mathbf{q}$. The best we can do is use $\mathbf{\tilde{q}}$ to compute erroneous versions of these matrices: $\mathbf{\tilde{Q}}, \mathbf{\tilde{W}}, \mathbf{\tilde{S}}$.

Before moving forward, we assume that the contribution of $\mathbf{S}$ is negligible and choose to ignore it. We provide an explanation for this in Appendix B, and we additionally verify in Section V that this is a reasonable simplification.

That said, our control objective is to minimize the error between the end effector of the out-of-view chain and a goal pose, which we represent as $\mathbf{v}_{n_b} \in \mathbb{R}^4$ in the ``linear+roll" subspace:
\begin{equation} \label{vnb}
    \mathbf{v}_{n_b} = \mathbf{P}_{p,w_3}
    \begin{bmatrix}
        \mathbf{R}_{n_b}^{c} & \boldsymbol{0} \\
        \boldsymbol{0} & \mathbf{R}_{n_b}^{c}
    \end{bmatrix}^{\top}
    \begin{bmatrix}
        \Delta\mathbf{p}\\
        \Delta\mathbf{w}
    \end{bmatrix}
\end{equation}
Here, $\Delta\mathbf{p}$ and $\Delta\mathbf{w}$ are the linear and axis-angle displacements from the goal pose to the end effector specified in the camera frame, $\mathbf{R}_{n_b}^{c}$ is the rotational matrix component of the homogeneous transform from link frame $n_b$ to the camera frame, and $\mathbf{P}_{p,w_3} \in \mathbb{R}^{4 \times 6}$ is the 6D-to-4D projection constructed by removing rows 4 and 5 from the $6\times6$ identity matrix. Note that $\mathbf{R}_{n_b}^{c}$ can be computed from the out-of-view portion of \eqref{og_kc}, so this transform is still well defined when lumped-error tracking is incorporated in the control loop.

Putting everything together, we use $\mathbf{\tilde{Q}}, \mathbf{\tilde{W}}$, and $\mathbf{v}_{n_b}$ to define the following resolved-rate control law.
\begin{equation}\label{ctrl_4dof}
    \mathbf{\dot{q}_{1:n_b}} = -\alpha(\mathbf{\tilde{Q}} \mathbf{\tilde{W}})^{\dagger} \mathbf{v}_{n_b}
\end{equation}
where $\alpha$ is a positive scalar gain and $(\cdot)^{\dagger}$ is the pseudoinverse operator.

Even though $\mathbf{\tilde{Q}}, \mathbf{\tilde{W}}$ are erroneous, we find that \eqref{ctrl_4dof} is provably stable. Our proof is detailed in Appendix A and holds under the following reasonable assumptions:

\begin{itemize}
    \item \textbf{Assumption 1:} $\|\dot{q}_1 \sin q_2 \| \ll \|\dot{q}_4\|$
    \item \textbf{Assumption 2:} $\mathbf{\dot{e}_{1:n_b}}$ is negligible
    \item \textbf{Assumption 3:} $|e_{n_b}| < \tau$, where $\tau$ is dependent on robot joint limits. In the case of the dVRK, $\tau = \frac{5\pi}{12}$ rad, or $75^{\circ}$
\end{itemize}
Justifications for these assumptions are provided in Appendix B.

Assumption 3 defines our bound on the set of out-of-view joint angle errors within which this controller is stable. This bound leaves $\mathbf{e_{1:3}}$ unconstrained, as long as $\mathbf{q}$ and $\mathbf{\tilde{q}}$ lie within the robot's joint limits. Furthermore, it's almost impossible to violate this bound in real-world settings, where $e_4$ is only around $10^{\circ}$ on average \cite{hwang2020efficiently}.

\subsection{Full Chain Control}

Because the out-of-view chain has only 4 degrees of freedom, it can only achieve pose goals in a 4D subspace of the robot workspace, which can have up to 6 dimensions. Hence, we need a full chain controller to achieve all poses in the robot workspace. We can integrate our out-of-view controller into a bilevel control scheme for the full chain.

First, we define our high-level control objective as follows:
\begin{equation} \label{vn}
    \mathbf{v}_{n} =
    \begin{bmatrix}
        \mathbf{R} & \boldsymbol{0} \\
        \boldsymbol{0} & \mathbf{R}
    \end{bmatrix}^{\top}
    \begin{bmatrix}
        \Delta\mathbf{p}\\
        \Delta\mathbf{w}
    \end{bmatrix}
\end{equation}
where $\mathbf{R}$ is the rotation matrix component of \eqref{og_kc}. The high-level controller takes actions by assigning low-level control objectives to the out-of-view chain and in-view chain. The low-level control objective for the out-of-view chain will be some $\mathbf{\hat{v}}_{n_b} \in \mathbb{R}^4$. We define $\mathbf{\hat{v}}_{n}^{n_b} \in \mathbb{R}^{6}$ as the low-level control objective for the in-view chain, computed by taking the difference between $\mathbf{\hat{v}}_n$ and the estimated twist impact of $\mathbf{v}_{n_b}$ on link $n$:
\begin{equation} \label{vn_nb}
    \mathbf{\hat{v}}_{n}^{n_b} = \mathbf{v}_n - [\mathbf{Ad}_{n_b}^{n}]\mathbf{\tilde{D}}\mathbf{\hat{v}}_{n_b}
\end{equation}
Here, $[\mathbf{Ad}_{n_b}^{n}]$ is the adjoint twist transform from frame $n_b$ to $n$, which is known because these link frames lie in-view. This way, the end-effector twist goals associated with $\mathbf{\hat{v}}_n$ and $\mathbf{\hat{v}}_{n}^{n_b}$ add up to $\mathbf{v}_n$.

Before defining our high-level control law, we must first choose a controller for the in-view chain. We select the following resolved-rate controller.
\begin{equation} \label{ctrl_inview}
    \mathbf{\dot{q}_{n_b+1:n}} = -\alpha(\mathbf{J}_{n}^{n_b})^{\dagger} \mathbf{\hat{v}}_n^{n_b}
\end{equation}
Note that $\mathbf{J}_{n}^{n_b}$ is the true body-frame Jacobian of the in-view chain, which is computable because $\mathbf{q}_{n_b+1:n}$ is known. Because \eqref{ctrl_inview} uses the pseudoinverse of this true Jacobian, it is guaranteed to be stable \cite{wampler1988applications}. However, it's important to note that for arbitrary $\mathbf{\hat{v}}_n^{n_b}$ the computed joint velocities will produce $\mathbf{J}_{n}^{n_b}(\mathbf{J}_{n}^{n_b})^{\dagger}\mathbf{\hat{v}}_{n}^{n_b}$, which is the orthogonal projection of $\mathbf{\hat{v}}_n^{n_b}$ into the column space of $\mathbf{J}_{n}^{n_b}$. If the in-view chain is underactuated, as it often is, then this column space will be an $n - n_b$ dimensional subspace of $\mathbb{R}^6$. The high-level controller must take the impact of this projection into account when selecting $\mathbf{\hat{v}}_n^{n_b}$.

With both low-level controllers defined, we can now define the high-level control law as the following closed-form least squares optimization.
\begin{equation} \label{ctrl_6dof}
\begin{aligned}
    \argmin_{\mathbf{\hat{v}}_{n_b} \in \mathbb{R}^4} & \| \mathbf{v}_n - [\mathbf{Ad}_{n_b}^{n}]\mathbf{\tilde{D}}\mathbf{\hat{v}}_{n_b} - \mathbf{J}_{n}^{n_b}(\mathbf{J}_{n}^{n_b})^{\dagger}\mathbf{\hat{v}}_{n}^{n_b}\|\\
    &\textbf{subject to} \textrm{ \eqref{vn_nb}}
\end{aligned}
\end{equation}
Note that the high-level controller need only select $\mathbf{\hat{v}}_{n_b}$, as $\mathbf{\hat{v}}_{n}^{n_b}$ can be computed afterwards from \eqref{vn_nb}. Intutively, this controller coordinates the out-of-view and in-view controllers such that their combined estimated twist output is $-\alpha\mathbf{v}_n$, taking into account the degrees of freedom of the in-view chain.

Unlike the IK controller, both our resolved-rate controllers are non-iterative and closely grounded in the stability of \eqref{ctrl_4dof} and \eqref{ctrl_inview}. Furthermore, as shown in Section V, this ensures it can handle a wider range of tool configurations.

\section{EXPERIMENTS \& RESULTS}

\begin{figure*}[t]
    \centering
    \vspace{5mm}\includegraphics[width=\linewidth]{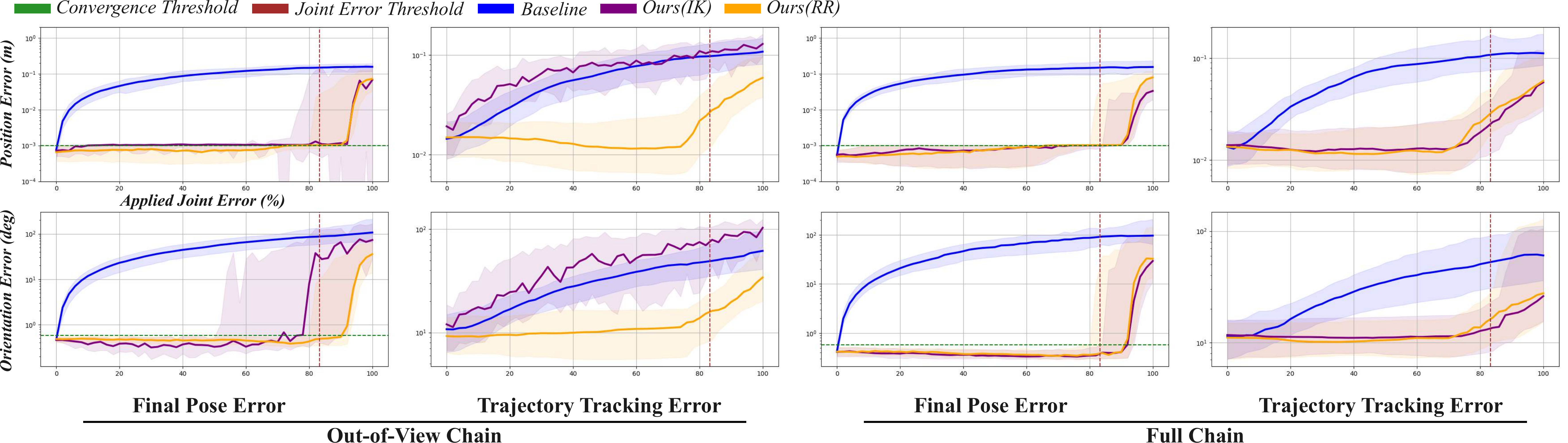}
    \caption{Simulated experiment results for out-of-view chain (left) and full chain (right) control. We gradually added errors to the out-of-view joints and tracked how final pose error and trajectory tracking error degraded, plotting the median and inter-quartile range of these metrics. The calibration-based baseline rapidly fails in both cases. Our previous IK controller struggles to control the underactuated out-of-view chain because of workspace mismatch, but performs identically to our resolved-rate (RR) method when controlling the full chain. Our resolved-rate controller performs well in both cases, typically failing only after the error applied exceeds the marked theoretical threshold.}
    \label{sim_exp}
\end{figure*}

\begin{table*}[t]
    \centering
    \caption{Results from Out-of-View Chain Simulated Experiments}
    \label{sim_table_oov}
    \begin{tabular}{c|c|ccccc|ccccc}
        \toprule
        \multirow{2}{*}{\centering Method} & \multirow{2}{1cm}{\centering Metric} & \multicolumn{5}{c|}{$\downarrow$ Final Pose Error (Median$\pm$IQR/2)} & \multicolumn{5}{c}{$\downarrow$ Tracking Error (Median$\pm$IQR/2)}\\
        && 0\% & 33\% & 66\% & ~83.3\% & 100\% & 0\% & 33\% & 66\%  & ~83.3\% & 100\% \\
        \hline
        \multirow{2}{*}{\centering Baseline} & Pos.(mm) & 0.8$\pm$0.3 & 70.3$\pm$26.3 & 127.5$\pm$35.7 & 146.9$\pm$39.1  & 156.2$\pm$38.9 & \textbf{14.4$\pm$6.2} & 46.8$\pm$14.6 & 83.8$\pm$26.7 & 96.7$\pm$31.9  & 108.0$\pm$33.5 \\ 
        & Ori.(deg) & \textbf{0.5$\pm$0.1} & 36.2$\pm$12.1 & 71.2$\pm$22.6 & 86.1$\pm$37.1  & 106.7$\pm$71.9 & 10.8$\pm$4.2 & 23.5$\pm$6.5 & 42.3$\pm$11.9 & 48.4$\pm$17.9  & 61.7$\pm$26.6 \\
        \hline
        \multirow{2}{*}{\centering Ours(IK)} & Pos.(mm) & 0.7$\pm$0.3 & 1.0$\pm$0.2 & 1.0$\pm$0.3 & 1.3$\pm$62.2  & \textbf{67.2$\pm$81.7} & 19.2$\pm$7.8 & 72.6$\pm$24.7 & 80.6$\pm$37.1 & 103.8$\pm$34.9  & 129.1$\pm$32.5 \\ 
        & Ori.(deg) & \textbf{0.5$\pm$0.1} & \textbf{0.3$\pm$0.2} & 0.4$\pm$14.9 & 37.5$\pm$65.3  & 73.0$\pm$68.6 & 12.0$\pm$5.9 & 43.2$\pm$25.4 & 56.5$\pm$33.4 & 69.9$\pm$35.3  & 103.3$\pm$27.2 \\
        \hline
        \multirow{2}{*}{\centering Ours(RR)} & Pos.(mm) & \textbf{0.6$\pm$0.3} & \textbf{0.8$\pm$0.4} & \textbf{0.8$\pm$0.3} & \textbf{1.0$\pm$10.5}  & 72.1$\pm$49.0 & 14.9$\pm$5.9 & \textbf{13.8$\pm$6.9} & \textbf{11.5$\pm$8.7} & \textbf{24.3$\pm$16.4}  & \textbf{59.0$\pm$29.7} \\ 
        & Ori.(deg) & \textbf{0.5$\pm$0.1} & 0.4$\pm$0.1 & \textbf{0.4$\pm$0.1} & \textbf{0.5$\pm$3.8}  & \textbf{35.7$\pm$25.6} & \textbf{9.2$\pm$3.9} & \textbf{9.8$\pm$5.0} & \textbf{10.9$\pm$5.5} & \textbf{15.3$\pm$7.5}  & \textbf{34.1$\pm$17.9} \\
        \bottomrule
    \end{tabular}
\end{table*}

\begin{table*}[t]
    \centering
    \caption{Results from Full Chain Simulated Experiments}
    \label{sim_table_full}
    \begin{tabular}{c|c|ccccc|ccccc}
        \toprule
        \multirow{2}{*}{\centering Method} & \multirow{2}{1cm}{\centering Metric} & \multicolumn{5}{c|}{$\downarrow$ Final Pose Error (Median$\pm$IQR/2)} & \multicolumn{5}{c}{$\downarrow$ Tracking Error (Median$\pm$IQR/2)}\\
        && 0\% & 33\% & 66\% & ~83.3\% & 100\% & 0\% & 33\% & 66\%  & ~83.3\% & 100\% \\
        \hline
        \multirow{2}{*}{\centering Baseline} & Pos.(mm) & \textbf{0.5$\pm$0.2} & 79.0$\pm$28.6 & 134.8$\pm$46.2 & 146.7$\pm$62.1  & 154.2$\pm$60.1 & \textbf{13.4$\pm$5.0} & 51.5$\pm$13.8 & 91.7$\pm$34.8 & 106.8$\pm$44.4  & 111.6$\pm$43.5 \\ 
        & Ori.(deg) & \textbf{0.4$\pm$0.1} & 34.2$\pm$13.9 & 72.1$\pm$27.8 & 89.1$\pm$39.6  & 96.2$\pm$72.0 & 11.4$\pm$4.4 & 23.7$\pm$8.4 & 43.2$\pm$22.3 & 51.8$\pm$29.5  & 60.5$\pm$37.6 \\
        \hline
        \multirow{2}{*}{\centering Ours(IK)} & Pos.(mm) & 0.6$\pm$0.2 & 0.7$\pm$0.4 & \textbf{0.9$\pm$0.3} & \textbf{1.0$\pm$0.1}  & \textbf{33.2$\pm$37.6} & 13.9$\pm$5.3 & 12.6$\pm$6.0 & 12.9$\pm$8.0 & \textbf{20.8$\pm$13.2}  & \textbf{57.8$\pm$43.5} \\ 
        & Ori.(deg) & \textbf{0.4$\pm$0.1} & \textbf{0.4$\pm$0.1} & \textbf{0.3$\pm$0.1} & \textbf{0.4$\pm$0.1}  & \textbf{29.4$\pm$42.4} & 11.7$\pm$4.4 & 11.1$\pm$4.3 & 11.4$\pm$5.0 & \textbf{13.1$\pm$6.8}  & \textbf{26.1$\pm$45.0} \\
        \hline
        \multirow{2}{*}{\centering Ours(RR)} & Pos.(mm) & \textbf{0.5$\pm$0.2} & \textbf{0.6$\pm$0.4} & 1.0$\pm$0.3 & 1.0$\pm$10.3  & 78.6$\pm$42.1 & 13.5$\pm$4.7 & \textbf{11.7$\pm$5.6} & \textbf{11.9$\pm$7.9} & 26.3$\pm$15.9  & 60.0$\pm$38.1 \\ 
        & Ori.(deg) & \textbf{0.4$\pm$0.1} & \textbf{0.4$\pm$0.1} & 0.4$\pm$0.1 & 0.4$\pm$5.0  & 32.5$\pm$65.2 & \textbf{11.1$\pm$4.0} & \textbf{10.1$\pm$3.8} & \textbf{11.0$\pm$4.3} & 15.4$\pm$8.1  & 27.6$\pm$56.3 \\
        \bottomrule
    \end{tabular}
\end{table*}

We implement our controllers in both simulation and hardware to verify our analytical results. We used Pytorch Kinematics \cite{Zhong_PyTorch_Kinematics_2024} for all kinematics-related computation. We specifically tested on the dVRK with Large Needle Driver (LND) attachments for both simulation and hardware.

\subsection{Simulated Experiments}

We implemented a kinematics-based simulator and rolled out several control trajectories, examining how performance degrades as we proportionally increase out-of-view joint angle errors. All results are visualized in Fig. \ref{sim_exp}, with selected results additionally detailed in Fig. \ref{sim_examples}, Table \ref{sim_table_oov}, and Table \ref{sim_table_full}. We test both our previously-developed IK-based controller and our novel resolved-rate controller, along with a baseline resolved-rate controller that uses an initial calibration rather than lumped error tracking. We conducted 2 sets of experiments, one where we control only the out-of-view chain (i.e. endo-wrist excluded) and another where we control the full chain (i.e. endo-wrist included). In each set, we randomly sample 50 joint angle configurations and use forward kinematics (FK) to produce 50 start and end poses that are guaranteed to lie within the current chain's workspace. Each simulated episode is 600 iterations, split into 2 stages: in the first 60 iterations, we define a goal trajectory between the start and end poses that the controller must follow, visualized in Fig. \ref{sim_examples}; in the remaining 540 iterations, the goal remains stationary at the end pose. We re-run this 50-trajectory benchmark 51 times, where for each test run $i$ we apply the error bias $\mathbf{e_{1:n_b}} = \pm \frac{i-1}{50}\mathbf{e_{1:n_b}^{max}}$ to produce erroneous joint angle readings in the out-of-view joints. We set $\mathbf{e_{1:n_b}^{max}} = [0^{\circ}  ~53^{\circ} ~0.1835\textrm{m} ~90^{\circ}]^{\top}$ to ensure test $i=50$ applies worst-case joint angle errors. We ignore outer yaw (joint 1) error as this has no impact on control performance. We measure the final pose error and root mean square tracking error for each simulated rollout, as visualized in Fig. \ref{sim_exp}. Additional simulator details are as follows: for each simulation step, we apply control velocities for a finite time step using FK; we set $\alpha = \frac{1}{6}$ as the gain for all controllers; lastly, if the end effector ever reaches within $1$ mm and $0.5$ deg of the final goal pose, we consider it ``converged" and end the rollout early.

We find that our resolved-rate controller's behavior closely matches our theoretical expectations. As shown in the final pose error graphs in Fig. \ref{sim_exp}, it tends to converge as long as the applied error is less than $83.3\%$, at which $\|e_4\|$ hits our bound of $75^{\circ}$. This behavior is well visualized in Fig. \ref{sim_examples}, where our controller's path begins to oscillate around the goal until it eventually fails to converge when $\|e_4\|$ grows too large. Note that this result supports the simplifications we make in Section IV-A, particularly the omission of $\mathbf{S}$. Additionally, our method easily outperforms the baseline, highlighting the advantage of including visual tracking in the loop versus relying on a fragile calibration.

We glean additional insights by comparing our IK-based controller with our resolved-rate controller. In the out-of-view chain case, the IK-based controller struggles. This is because the erroneous out-of-view chain has a different workspace than the true out-of-view chain. As a result, although the goal poses lay in the workspace of the true out-of-view chain, the IK solver often couldn't converge to a valid solution, resulting in unstable behavior. This problem only arises for underactuated chains, which is why the IK controller peforms well in the full chain case. Here, the performances of our IK and resolved-rate controllers are almost identical, suggesting that our theoretical analysis may be applicable to a wider range of kinematics-based controllers. Overall, our resolved-rate controller is the best option for underactuated chains, while both our controllers can be used interchangeably for fully/over-actuated chains.

\subsection{Real World Experiments}

\begin{figure}[t]
    \centering
    \vspace{5mm}\includegraphics[width=\linewidth]{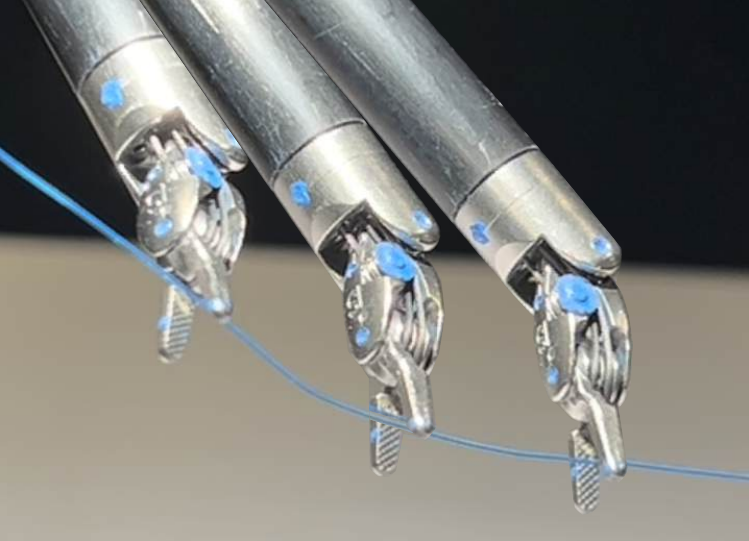}
    \caption{We implement our controller on hardware and verify its performance with suture thread grasping trials.}
    \label{real_exp}
\end{figure}

\begin{table}[t]
\vspace{5mm}
\caption{Suture Thread Grasp Success Rate}
\label{real_table}
\begin{center}
\begin{tabular}{c|ccc}
\toprule
 & Baseline & Ours (IK) & Ours (RR)\\
\hline
Success Rate & 0/20 & 19/20 & 17/20\\
\bottomrule
\end{tabular}
\end{center}
\end{table}

As shown in Fig. \ref{real_exp}, we conduct suture thread grasping experiments to assess our controller's performance on hardware. Our results are summarized in Table \ref{real_table}. We used the suture thread reconstruction method developed in \cite{joglekar2024autonomous}. We selected a thread configuration with minimal epipolar alignment to ensure reliable reconstruction. We used the particle filter developed in \cite{richter2021robotic} to perform tool tracking. Each controller was tasked with grasping 20 randomly-ordered points along the thread. We only allowed the gripper to open to a max angle of $20^{\circ}$ before grasping, increasing the precision required to complete the task.

As shown in Table \ref{real_table}, both of our controllers performed nearly identically with high accuracy. The discrepancy in final grasp success rate is mainly due to intermittent tracking errors that happened to appear more often in the resolved-rate (RR) experiment. Meanwhile, the baseline narrowly misses every time because of initial calibration error.

\section{DISCUSSION \& CONCLUSION}

We design a provably stable tracking-in-the-loop resolved-rate controller for the out-of-view portion of the RCM kinematic chain. We specify a bound on the out-of-view joint angle errors within which control stability holds, and our resolved-rate controller's convergence behavior closely adheres to this bound in our experiments. We additionally use a bilevel scheme to apply this controller to full chain control, agnostic to the type of tool attachment.

Despite its relatively simple formulation, our resolved-rate controller performs better than our previous IK controller for underactuated manipulators. As a result, it's likely more applicable for underactuated tool configurations, such as blood suction tools which only have 5 degrees of freedom. Nevertheless, our full chain IK controller behaves almost identically to our full chain resolved-rate controller, suggesting that our theoretical joint angle error bound may be applicable to a wider range of kinematics-based RCM controllers.

Although our case study mainly focuses on the dVRK, our theoretical results apply to any serial RCM chain with a similar joint order including virtually defined joints \cite{sandoval2017new}.
The exact joint error bound would vary depending on the manipulator joint limits, but can be derived in a similar manner to our work.
However, our analysis does not apply to controlling out-of-view joints beyond the RCM.
For example, if part of the LND endo-wrist were moved out of the endoscopic view then the out-of-view chain may become too complex to ensure stable control.
Instead, our work emphasizes the advantage of current serial RCM manipulator and tool attachment designs.

We provide key insight into improving the reliability of downstream surgical autonomy tasks.
Furthermore, we verify that lumped error tracking remains powerful framework for partially-observable settings.
As noted in our suture thread grasping experiments, currently the biggest bottleneck to performance is tracking, which relied on painted blue markers.
In future work, we will incorporate more consistently detected features for more accurate and precise tracking \cite{liang2025differentiable}.
Nevertheless, our lumped error controller will continue to enable high-precision downstream tasks like suture needle picking \cite{ho2024surgirl}, needle handoff \cite{chiu2021bimanual}, and suture thread manipulation \cite{joglekar2024autonomous}, promoting the realization of safe surgical autonomy.



\section*{APPENDIX}




\subsection{Proof of Out-of-View Control Stability}

\textbf{Claim:} The control law defined by \eqref{ctrl_4dof} is stable under Assumptions 1-3

\textbf{Proof:} Let $\boldsymbol{v}_{n_b}$ be the velocity of link $n_b$. Under Assumptions 1 and 2, we can write $\boldsymbol{v}_{n_b}$ based on \eqref{ctrl_4dof} as follows:
\begin{equation} \label{v_proof}
    \boldsymbol{v}_{n_b} = -\alpha\mathbf{Q}\mathbf{W}(\mathbf{\tilde{Q}} \mathbf{\tilde{W}})^{\dagger} \mathbf{v}_{n_b}
\end{equation}

We proceed with Lyapunov stability analysis. Let $V = \frac{1}{2}\mathbf{v}_{n_b}^{\top}\mathbf{v}_{n_b}$ be our Lyapunov potential function. It's derivative is $\dot{V} = \mathbf{v}_{n}^{\top}\boldsymbol{v}_{n_b}$. Plugging in \eqref{v_proof} and rearranging terms, we can rewrite $\dot{V}$ as follows.
\begin{equation} \label{Vdot}
\begin{aligned}[b]
    \dot{V} &= -\alpha\mathbf{v}_{n_b}^{\top}\mathbf{\tilde{Q}}^{\top} \mathbf{M} \mathbf{\tilde{Q}}\mathbf{v}_{n_b}\\
    \mathbf{M} &= \begin{bmatrix}
        \frac{q_3\cos q_2}{\tilde{q}_3\cos \tilde{q}_2}\cos e_4  & \frac{q_3}{\tilde{q}_3}\sin e_4  & 0 & 0\\
        -\frac{q_3\cos q_2 }{\tilde{q}_3\cos \tilde{q}_2}\sin e_4  & \frac{q_3}{\tilde{q}_3}\cos e_4  & 0 & 0\\
        0 & 0 & 1 & 0\\
        0 & 0 & 0 & 1
    \end{bmatrix}
\end{aligned}
\end{equation}

Our system is stable if $\dot{V} < 0$. This is true if $\mathbf{x}^{\top}\mathbf{M}\mathbf{x} > 0$ for arbitrary $\mathbf{x} \in \mathbb{R}^4$ when $\mathbf{x} \neq \boldsymbol{0}$. Expanded into its quadratic form, $\mathbf{x}^{\top}\mathbf{M}\mathbf{x}$ is a 4D paraboloid with the following Hessian.
\begin{equation} \label{H}
    \mathbf{H} = \begin{bmatrix}
        2\frac{q_3\cos q_2 }{\tilde{q}_3\cos \tilde{q}_2 }\cos e_4 & \frac{q_3}{\tilde{q}_3}(1-\frac{\cos q_2}{\cos \tilde{q}_2})\sin e_4 & 0 & 0\\
        \frac{q_3}{\tilde{q}_3}(1-\frac{\cos q_2}{\cos \tilde{q}_2})\sin e_4 & 2\frac{q_3}{\tilde{q}_3}\cos e_4 & 0 & 0\\
        0 & 0 & 2 & 0\\
        0 & 0 & 0 & 2
    \end{bmatrix}
\end{equation}
Applying Sylvester's Condition \cite{gilbert1991positive}, we find that $\mathbf{H}$ is positive definite within the joint error range specified by Assumption 3. This means that $\mathbf{x}^{\top}\mathbf{M}\mathbf{x}$ has a global minimum at $\mathbf{x} = \boldsymbol{0}$, so our system is globally asymptotically stable $\blacksquare$

\subsection{Explanation of Assumptions}

\textbf{Assumption 1:} The term $\dot{q}_1 \sin q_2$ arises from $\mathbf{S}$ and describes how the outer yaw joint $q_1$
can impact roll velocity in the link $n_b$ frame. This term tends to be small because $\sin q_2$ is often small, and in practice it can be ignored because roll joint $q_4$ can easily correct for its impact. Hence, we can safely assume $\|\dot{q}_1 \sin q_2 \| \ll \|\dot{q}_4\|$.

\textbf{Assumption 2:} We seek to analyze how kinematic chain errors, manifested as joint error biases, impact control stability. Hence, considering $\mathbf{\dot{e}_{1:n_b}}$ lies out of scope of our work. Furthermore, as verified in our real world experiments, $\mathbf{\dot{e}_{1:n_b}}$ is small enough to be ignored in practice.

\textbf{Assumption 3:} This assumption serves as our definition of the joint angle error bound $|e_{n_b}| < \tau$, within which our controller remains stable. Note that the robot joint limits implicitly bound the remaining 3 out-of-view joints.

\bibliographystyle{IEEEtran}
\bibliography{root}

\begin{thebibliography}{10}
\providecommand{\url}[1]{#1}
\csname url@samestyle\endcsname
\providecommand{\newblock}{\relax}
\providecommand{\bibinfo}[2]{#2}
\providecommand{\BIBentrySTDinterwordspacing}{\spaceskip=0pt\relax}
\providecommand{\BIBentryALTinterwordstretchfactor}{4}
\providecommand{\BIBentryALTinterwordspacing}{\spaceskip=\fontdimen2\font plus
\BIBentryALTinterwordstretchfactor\fontdimen3\font minus \fontdimen4\font\relax}
\providecommand{\BIBforeignlanguage}[2]{{%
\expandafter\ifx\csname l@#1\endcsname\relax
\typeout{** WARNING: IEEEtran.bst: No hyphenation pattern has been}%
\typeout{** loaded for the language `#1'. Using the pattern for}%
\typeout{** the default language instead.}%
\else
\language=\csname l@#1\endcsname
\fi
#2}}
\providecommand{\BIBdecl}{\relax}
\BIBdecl

\bibitem{zhang2024state}
W.~Zhang, Z.~Wang, K.~Ma, F.~Liu, P.~Cheng, and X.~Ding, ``State of the art in movement around a remote point: a review of remote center of motion in robotics,'' \emph{Frontiers of Mechanical Engineering}, vol.~19, no.~2, p.~14, 2024.

\bibitem{ostrander2024current}
B.~T. Ostrander, D.~Massillon, L.~Meller, Z.-Y. Chiu, M.~Yip, and R.~K. Orosco, ``The current state of autonomous suturing: a systematic review,'' \emph{Surgical Endoscopy}, vol.~38, no.~5, pp. 2383--2397, 2024.

\bibitem{hari2024stitch}
K.~Hari, H.~Kim, W.~Panitch, K.~Srinivas, V.~Schorp, K.~Dharmarajan, S.~Ganti, T.~Sadjadpour, and K.~Goldberg, ``Stitch: Augmented dexterity for suture throws including thread coordination and handoffs,'' in \emph{2024 International Symposium on Medical Robotics (ISMR)}.\hskip 1em plus 0.5em minus 0.4em\relax IEEE, 2024, pp. 1--7.

\bibitem{ge2023autonomous}
J.~Ge, M.~Kam, J.~D. Opfermann, H.~Saeidi, S.~Leonard, L.~J. Mady, M.~J. Schnermann, and A.~Krieger, ``Autonomous system for tumor resection (astr)-dual-arm robotic midline partial glossectomy,'' \emph{IEEE robotics and automation letters}, vol.~9, no.~2, pp. 1166--1173, 2023.

\bibitem{liang2024medic}
X.~Liang, C.-P. Wang, N.~U. Shinde, F.~Liu, F.~Richter, and M.~Yip, ``Medic: Autonomous surgical robotic assistance to maximizing exposure for dissection and cautery,'' \emph{arXiv preprint arXiv:2409.14287}, 2024.

\bibitem{Fei_2021_R2S}
F.~Liu, Z.~Li, Y.~Han, J.~Lu, F.~Richter, and M.~C. Yip, ``Real-to-sim registration of deformable soft tissue with position-based dynamics for surgical robot autonomy,'' in \emph{2021 IEEE International Conference on Robotics and Automation (ICRA)}, 2021, pp. 12\,328--12\,334.

\bibitem{richter2021autonomous}
F.~Richter, S.~Shen, F.~Liu, J.~Huang, E.~K. Funk, R.~K. Orosco, and M.~C. Yip, ``Autonomous robotic suction to clear the surgical field for hemostasis using image-based blood flow detection,'' \emph{IEEE Robotics and Automation Letters}, vol.~6, no.~2, pp. 1383--1390, 2021.

\bibitem{Jingbin_2021_blood}
J.~Huang, F.~Liu, F.~Richter, and M.~C. Yip, ``Model-predictive control of blood suction for surgical hemostasis using differentiable fluid simulations,'' in \emph{2021 IEEE International Conference on Robotics and Automation (ICRA)}, 2021, pp. 12\,380--12\,386.

\bibitem{ou2024autonomous}
Y.~Ou, A.~Soleymani, X.~Li, and M.~Tavakoli, ``Autonomous blood suction for robot-assisted surgery: A sim-to-real reinforcement learning approach,'' \emph{IEEE Robotics and Automation Letters}, 2024.

\bibitem{shinde2024surestep}
N.~U. Shinde, Z.-Y. Chiu, F.~Richter, J.~Lim, Y.~Zhi, S.~Herbert, and M.~C. Yip, ``Surestep: An uncertainty-aware trajectory optimization framework to enhance visual tool tracking for robust surgical automation,'' in \emph{2024 IEEE/RSJ International Conference on Intelligent Robots and Systems (IROS)}.\hskip 1em plus 0.5em minus 0.4em\relax IEEE, 2024, pp. 6953--6960.

\bibitem{joglekar2024autonomous}
N.~Joglekar, F.~Liu, F.~Richter, and M.~C. Yip, ``Autonomous image-to-grasp robotic suturing using reliability-driven suture thread reconstruction,'' \emph{IEEE Robotics and Automation Letters}, vol.~10, no.~4, pp. 3676--3683, 2025.

\bibitem{shinde2024jiggle}
N.~U. Shinde, X.~Liang, F.~Liu, Y.~Zhang, F.~Richter, S.~Herbert, and M.~C. Yip, ``Jiggle: An active sensing framework for boundary parameters estimation in deformable surgical environments,'' \emph{arXiv preprint arXiv:2405.09743}, 2024.

\bibitem{blausen2014medical}
B.~Medical, ``Medical gallery of blausen medical 2014,'' \emph{WikiJournal of Medicine}, vol.~1, no.~2, pp. 1--79, 2014.

\bibitem{richter2021robotic}
F.~Richter, J.~Lu, R.~K. Orosco, and M.~C. Yip, ``Robotic tool tracking under partially visible kinematic chain: A unified approach,'' \emph{IEEE Transactions on Robotics}, vol.~38, no.~3, pp. 1653--1670, 2021.

\bibitem{peng2020real}
H.~Peng, X.~Yang, Y.-H. Su, and B.~Hannaford, ``Real-time data driven precision estimator for raven-ii surgical robot end effector position,'' in \emph{2020 IEEE International Conference on Robotics and Automation (ICRA)}.\hskip 1em plus 0.5em minus 0.4em\relax IEEE, 2020, pp. 350--356.

\bibitem{hwang2020efficiently}
M.~Hwang, B.~Thananjeyan, S.~Paradis, D.~Seita, J.~Ichnowski, D.~Fer, T.~Low, and K.~Goldberg, ``Efficiently calibrating cable-driven surgical robots with rgbd fiducial sensing and recurrent neural networks,'' \emph{IEEE Robotics and Automation Letters}, vol.~5, no.~4, pp. 5937--5944, 2020.

\bibitem{ye2016real}
M.~Ye, L.~Zhang, S.~Giannarou, and G.-Z. Yang, ``Real-time 3d tracking of articulated tools for robotic surgery,'' in \emph{Medical Image Computing and Computer-Assisted Intervention--MICCAI 2016: 19th International Conference, Athens, Greece, October 17-21, 2016, Proceedings, Part I 19}.\hskip 1em plus 0.5em minus 0.4em\relax Springer, 2016, pp. 386--394.

\bibitem{hao2018vision}
R.~Hao, O.~Özgüner, and M.~C. Çavuşoğlu, ``Vision-based surgical tool pose estimation for the da vinci® robotic surgical system,'' in \emph{2018 IEEE/RSJ International Conference on Intelligent Robots and Systems (IROS)}, 2018, pp. 1298--1305.

\bibitem{reiter2014appearance}
A.~Reiter, P.~K. Allen, and T.~Zhao, ``Appearance learning for 3d tracking of robotic surgical tools,'' \emph{The International Journal of Robotics Research}, vol.~33, no.~2, pp. 342--356, 2014.

\bibitem{zhao2015efficient}
T.~Zhao, W.~Zhao, B.~D. Hoffman, W.~C. Nowlin, and H.~Hui, ``Efficient vision and kinematic data fusion for robotic surgical instruments and other applications,'' U.S. Patent 8,971,597, Mar. 3, 2015.

\bibitem{dambrosia2024robust}
C.~D’Ambrosia, F.~Richter, Z.-Y. Chiu, N.~Shinde, F.~Liu, H.~I. Christensen, and M.~C. Yip, ``Robust surgical tool tracking with pixel-based probabilities for projected geometric primitives,'' in \emph{2024 IEEE International Conference on Robotics and Automation (ICRA)}, 2024, pp. 15\,455--15\,462.

\bibitem{chiu2021bimanual}
Z.-Y. Chiu, F.~Richter, E.~K. Funk, R.~K. Orosco, and M.~C. Yip, ``Bimanual regrasping for suture needles using reinforcement learning for rapid motion planning,'' in \emph{2021 IEEE International Conference on Robotics and Automation (ICRA)}.\hskip 1em plus 0.5em minus 0.4em\relax IEEE, 2021, pp. 7737--7743.

\bibitem{ho2024surgirl}
Y.-J. Ho, Z.-Y. Chiu, Y.~Zhi, and M.~C. Yip, ``Surgirl: Towards life-long learning for surgical automation by incremental reinforcement learning,'' \emph{arXiv preprint arXiv:2409.15651}, 2024.

\bibitem{wampler1988applications}
\BIBentryALTinterwordspacing
I.~Wampler, C.~W. and L.~J. Leifer, ``Applications of damped least-squares methods to resolved-rate and resolved-acceleration control of manipulators,'' \emph{Journal of Dynamic Systems, Measurement, and Control}, vol. 110, no.~1, pp. 31--38, 03 1988. [Online]. Available: \url{https://doi.org/10.1115/1.3152644}
\BIBentrySTDinterwordspacing

\bibitem{Zhong_PyTorch_Kinematics_2024}
S.~Zhong, T.~Power, A.~Gupta, and P.~Mitrano, ``{PyTorch Kinematics},'' Feb. 2024.

\bibitem{sandoval2017new}
J.~Sandoval, G.~Poisson, and P.~Vieyres, ``A new kinematic formulation of the rcm constraint for redundant torque-controlled robots,'' in \emph{2017 IEEE/RSJ International Conference on Intelligent Robots and Systems (IROS)}.\hskip 1em plus 0.5em minus 0.4em\relax IEEE, 2017, pp. 4576--4581.

\bibitem{liang2025differentiable}
Z.~Liang, Z.-Y. Chiu, F.~Richter, and M.~C. Yip, ``Differentiable rendering-based pose estimation for surgical robotic instruments,'' \emph{arXiv preprint arXiv:2503.05953}, 2025.

\bibitem{gilbert1991positive}
G.~T. Gilbert, ``Positive definite matrices and sylvester's criterion,'' \emph{The American Mathematical Monthly}, vol.~98, no.~1, pp. 44--46, 1991.

\end{thebibliography}
\balance

\end{document}